\documentclass[conference,compsoc]{IEEEtran}
\usepackage{amssymb,amsmath,epsfig,latexsym,graphicx,bm,multirow}
\usepackage{epstopdf}

\ifCLASSOPTIONcompsoc
  \usepackage[nocompress]{cite}
\else
  \usepackage{cite}
\fi
\ifCLASSINFOpdf
\else
\fi

\begin{document}
\title{Statistical Selection of CNN-Based Audiovisual Features for Instantaneous Estimation of Human Emotional States}


%\author{
%\IEEEauthorblockN{Ramesh Basnet\\ and Mohammad Tariqul Islam}
%\IEEEauthorblockA{Department of EEE, BUET\\Dhaka 1000, Bangladesh\\
%Email:rbasnet198@gmail.com\\tariqul@eee.buet.ac.bd}
%
%\and
%
%\IEEEauthorblockN{Tamanna Howlader} \IEEEauthorblockA{ISRT,
%University of Dhaka\\Dhaka 1000, Bangladesh\\
%Email:tamanna@isrt.ac.bd}
%
%\and
%
%\IEEEauthorblockN{S. M. Mahbubur Rahman}
%\IEEEauthorblockA{Department of EEE, ULAB \\ Dhaka 1209, Bangladesh\\
%Email: mahbubur.rahman@ulab.edu.bd}
%
%}


% compsoc conferences compared to traditional conferences), use this
\author{\IEEEauthorblockN{Ramesh Basnet\IEEEauthorrefmark{1},
Mohammad Tariqul Islam\IEEEauthorrefmark{1}, Tamanna
Howlader\IEEEauthorrefmark{2} \\ S. M. Mahbubur
Rahman\IEEEauthorrefmark{3}, and Dimitrios Hatzinakos
\IEEEauthorrefmark{4}}
\IEEEauthorblockA{\IEEEauthorrefmark{1}Department of EEE,
Bangladesh University of Engineering and Technology, Dhaka 1205,
Bangladesh}\IEEEauthorblockA{\IEEEauthorrefmark{2}Institute of
Statistical Research and Training, University of Dhaka, Dhaka
1000, Bangladesh}
\IEEEauthorblockA{\IEEEauthorrefmark{3}Department of EEE,
University of Liberal Arts Bangladesh, Dhaka 1209, Bangladesh}
\IEEEauthorblockA{\IEEEauthorrefmark{4} Department of ECE,
University of Toronto, Toronto, ON, Canada, M5S 2E4\\
Email: rbasnet198@gmail.com, tariqul@eee.buet.ac.bd,
tamanna@isrt.ac.bd\\mahbubur.rahman@ulab.edu.bd,
dimitris@comm.utoronto.ca}}


\maketitle

\begin{abstract}
Automatic prediction of continuous-level emotional state requires
selection of suitable affective features to develop a regression
system based on supervised machine learning. This paper
investigates the performance of features statistically learned
using convolutional neural networks for instantaneously predicting
the continuous dimensions of emotional states. Features with
minimum redundancy and maximum relevancy are chosen by using the
mutual information-based selection process. The performance of
frame-by-frame prediction of emotional state using the moderate
length features as proposed in this paper is evaluated on
spontaneous and naturalistic human-human conversation of RECOLA
database. Experimental results show that the proposed model can be
used for instantaneous prediction of emotional state with an
accuracy higher than traditional audio or video features that are
used for affective computation.
\end{abstract}


% \fi

%\IEEEpeerreviewmaketitle

\section{Introduction}
Automatic prediction of instantaneous affective state is becoming
increasingly important in the recent years. Analysis of affective
content is an interdisciplinary field involving research areas
that includes computer vision, speech analysis, and psychology. To
relate between measurable low-level features with corresponding
affective state, certain models of emotion are required.
Psychologists have used two major approaches, viz., categorical
and dimensional to quantify the emotional
states~\cite{wang2015video}. According to the categorical
approach, the model of emotion was defined by Ekman, who grouped
emotional states into six basic categories including the
happiness, sadness, anger, disgust, fear, and surprise
(see~\cite{ekman2005book}). However, categorical approach fall
short in situation where a small number of discrete categories may
not reflect the complexity of human emotional states. In this
context, continuous emotional model reflect more subtle and
context specific emotions avoiding boundaries. As a result,
research in area of affective computing is shifting from
categorical approach to dimensional approach. In dimensional
approach, Wundt~\cite{ref:wundt} introduced 3D continuous space -
valence, arousal, and dominance to represent emotional states of
humans.
\begin{figure}[t]
\centering
\includegraphics[width=7cm,height=7cm]{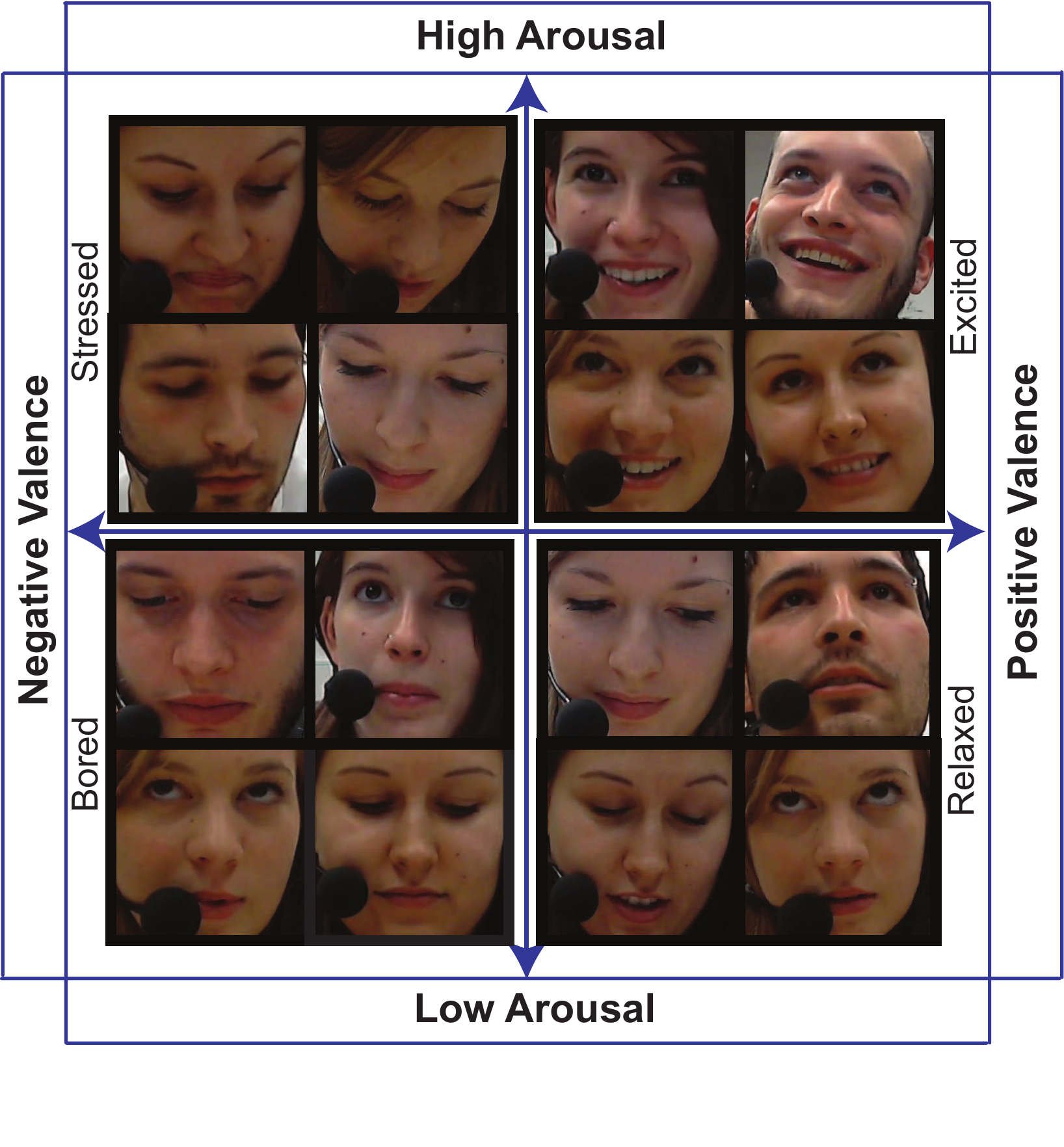}
\caption{Typical emotional states and human faces in the
valence-arousal plane.}\label{fig:va_plane}
\end{figure}
Valence represents the degree of pleasure ranging from pleasant to
unpleasant feelings. Arousal illustrates the activation level
ranging from global feeling of dynamism to lethargy of an
individual. Dominance characterizes the range of emotion from
controlling sentiment to the controlled or submissive feelings. It
is reported in \cite{dietz1999affective} that the effect of the
dominance dimension becomes visible only at points with distinctly
high absolute valence values. In general, the valence and arousal
account for most of the independent variance in emotional
responses (see for example, \cite{greenwald1989affective}).
Fig.~\ref{fig:va_plane} shows a few examples of facial appearances
in the valance-arousal plane. It is seen from this figure that
different emotional states such as stressed, excited, bored or
relaxed feelings can be recognized independent of the subjects by
the ratings of the dimensions valance and arousal. In general, the
affective states of humans are estimated by extracting affective
features from suitable sensors, and then relating between the
low-level descriptors and high-level semantic meanings. Initial
researches on affective content analysis confined in recognition
of exaggerated expressions of prototypical emotions that are
recorded in constrained environments~\cite{chen2000emotional}.
Spontaneous expressions are also recognized using the
differentially expressive components of the facial images
represented by orthogonal 2D Gaussian-Hermite
moments~\cite{imran2016differential}. Nevertheless,
challenges appear to solve for the problem of recognizing
continuous level naturalistic emotions instantaneously those are
displayed in our day-to-day life.

\begin{figure*}[t]
\centering
\includegraphics[width=14cm,height=6cm]{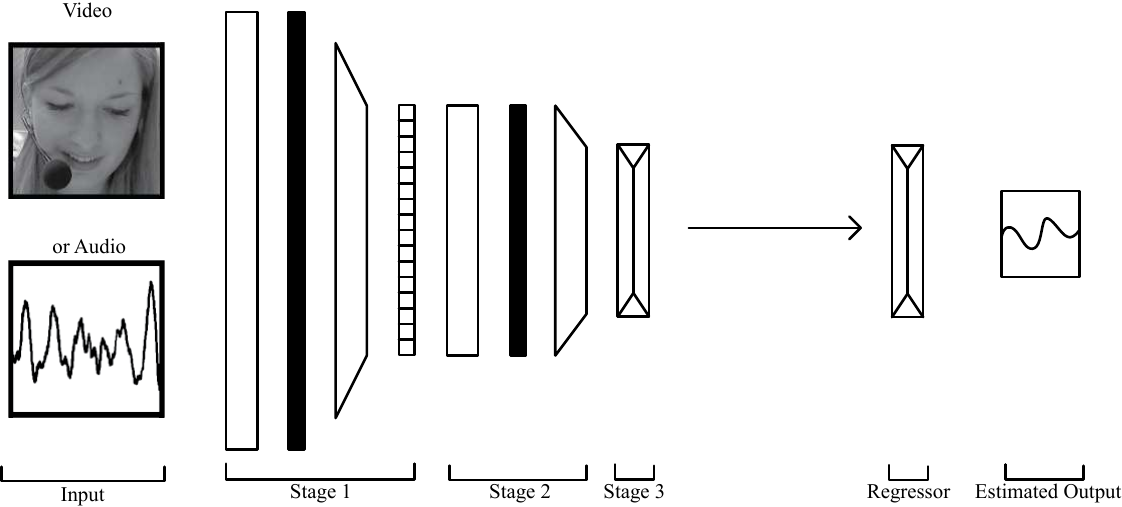}
\caption{Proposed CNN models for predicting the emotion states.
The input to the stream can be either video or audio signal.
Operations from left to right is considered the forward operation.
Each operations and layers of the network is represented by its
corresponding geometric shape. Convolution layer is given by a
rectangle, ReLU layer by a solid line, max-pooling layer by a
converging trapezoid, LRN layer by a striped rectangle, and fully
connected layer by an emerald shape.} \label{fig:models}
\end{figure*}

Prediction of continuous-level emotional states can be performed
by extracting features from audio, visual or physiological
signals. However, in practice, the former two-types of signals are
preferred to the last-type of signal for feature extraction,
because they are easily accessible and widely available. Finally,
the instantaneous prediction of emotion requires that the nature
of the features be dynamic instead of static. For example,
in~\cite{dahmane2011continuous}, the Gabor energy texture
descriptors of neighboring frames are used as features of emotion.
To remove redundant and irrelevant information in the regression
process, suitable selection techniques such as those employ the
correlation coefficients \cite{nicolle2012robust} and mutual
information~\cite{cui2013mutual} are applied to certain features.
Sudipta {\emph{et al.}}~\cite{sudipta2016affect} employed selected
set of audiovisual features from the mel frequency cepstral
coefficients (MFCC) and local binary patterns on three orthogonal
planes (LBP-TOP) in minimum redundancy and maximum relevance
(mRMR) pipeline to estimate continuous level emotional state in a
multimodal framework. Recently, machine learning algorithms such
as the hidden Markov model \cite{meng2011naturalistic}, the long
short-term memory (LSTM) network~\cite{wollmer2013lstm}, and the
convolutional neural network (CNN) with
LSTM~\cite{khorrami2016deep} have been applied to predict the
continuous-level emotional state.

% \subsection{Scope of Analysis}
Traditional methods of continuous-level affective computation use
a large number of hand-crafted features. In the literature, the
use of deep learning algorithms in affective computation is
relatively new. The most successful type of deep learning method
is the CNN~\cite{lecun2015deep}, which usually predict emotional
states by using a large set of affective features learned from the
training data. To the best of our knowledge, the outcome of
statistical selection of affective features learned by the CNN has
not yet been investigated. Thus, there remains a scope of
developing new deep learning algorithm using CNN such that the
frame-by-frame emotional state can be estimated in real-time by
the use of sufficiently low number of affective features chosen by
a suitable statistical selection process.
% \subsection{Specific Contributions}
In this paper, we predict the continuous-level human emotional
states instantaneously using a suitable selection of audiovisual
features learned by the CNN. Specifically, the frame-by-frame
instantaneous prediction of emotional states is performed by
adopting mutual information-based selection process of the
audiovisual features extracted from a suitable structure of CNN.
We show that improvement of prediction result is obtained when
only relevant features are taken care of and redundant features
are eliminated.

This paper is organized as follows. Features extracted from audio
and visual signals using a two-stream deep CNN are presented in
Section~2. The feature selection and regression processes are
detailed in Sections~3 and 4, respectively. Section~5 provides the
experimental setup and comparison of the results to evaluate the
proposed method. Finally, concluding remarks are given in
Section~6.

\section{Feature Extraction}
In order to extract relevant information from audio and video
signals for emotion prediction, we train an end-to-end CNN which
provides affective features.
%\subsection{CNN Architecture}
The proposed CNN architecture is a 3-stage filtering scheme that
employs the convolutional or fully connected layers as shown in
the stick diagram of Figure~\ref{fig:models}. In this network,
each of the convolution layers in the stages are followed by a
ReLU and a max-pool layer.  In the stick diagram of
Figure~\ref{fig:models}, the layers convolution, ReLU, max-pool
and LRN are shown using a rectangle, a solid line, a converging
trapezoid and a striped rectangle, respectively. The first stage
of the proposed CNN model uses the convolutional filtering with
ReLU activation and max-pool downsampling along with local
response normalization (LRN). The overall output for the input
frame $X_0$ in the first stage is represented as
\begin{align}
\mathbf{X_1}=LRN(MP(\max(0,\mathbf{X_0}\ast\mathbf{W_1}+\mathbf{b_1}
)))
\end{align}
where $\ast$ is the convolution operation, $\max(0,\cdot)$ is the
ReLU operation, $MP(\cdot)$ is the max-pool operation and
$LRN(\cdot)$ is the local response normalization operation. Second
stage uses the convolutional filtering with ReLU activation and
max-pool downsampling, which is given by
\begin{align}
\mathbf{X_2}=MP(\max(0,\mathbf{X_1}\ast\mathbf{W_2}+\mathbf{b_2}))
\end{align}
The final stage of the network is a fully connected layer given by
\begin{align}
\mathbf{X_3}=\mathbf{W_3^T}\mathbf{X_2}+\mathbf{b_3}
\end{align}
Overall in three stages, the parameters $\mathbf{W_1}$,
$\mathbf{W_2}$ are convolution filter sets with bias terms
$\mathbf{b_1}$ and $\mathbf{b_2}$, respectively, and
$\mathbf{W_3}$ is the weight matrix of the fully connected layer
with corresponding bias $\mathbf{b_3}$. We consider the output of
the third stage as the features learned by the network. The final
part of the CNN outputs the prediction of the emotional state, a
fully connected layer that results in only scalars given by
\begin{align}
y=\mathbf{W_r}^T\mathbf{X_3}+b_r
\end{align}
where $\mathbf{W_r}$ is the weight matrix, $\mathbf{b_r}$ is the
corresponding bias term of the regressor, and $y$ is the predicted
value of emotional dimension.

%\subsection{Training Scheme}
In order to learn the end-to-end mapping function for predicting
emotional state, the network parameters, i.e., the weights and
bias terms are required to be estimated. The mean squared error
(MSE) is easily differentiable and thus employed as the loss
function for mini-batch optimization. The gradient-based momentum
update algorithm~\cite{williams1986learning} is employed to
optimize the weights and bias terms. Moreover, the dropout
mechanism~\cite{srivastava2014dropout} is employed after third
stage of each of the networks for training purpose.
% Given, there are $N$
% frames in the mini-batch with estimated emotional sates $\mathbf{y}$
% and ground truth emotional states $\mathbf{y_{t}}$ for all the frames,
% then the MSE is given by
% \begin{align}
% &D = \frac{1}{N} \sum_{i=1}^{N} |y_{i}-y_{ti}|^2
% \end{align}
% Elements of the weights, denoted by $w$, are updated using the gradient of
% loss function $D$ with respect to $w$ using the gradient-based momentum
% update algorithm given by (\cite{williams1986learning})
% \begin{align}
% &\Delta w(\eta+1) = \mu \Delta w(\eta) - \lambda \frac{\partial D(\eta)}{\partial w(\eta)}\\
% &w(\eta+1) = w(\eta) + \Delta w(\eta+1)
% \end{align}
% where $\lambda$ $(\lambda>0)$ and $\mu$ $(\mu>0)$ are hyper-parameters
% known as learning rate and momentum term, respectively, and $\eta$ is
% the iteration number. In a similar fashion, the elements of bias terms
% are also updated.

% \subsection{Features}
We trained the proposed architecture of CNN for both the video and
audio streams. We add a superscripts $(v)$ for CNN trained on
video stream and $(a)$ for CNN trained on the audio stream. The
feature sets obtained from the video and audio streams using the
CNNs are denoted as $\mathbf{X_3^{(v)}}$ and $\mathbf{X_3^{(a)}}$,
respectively. These two sets are concatenated to obtain the
complete audiovisual feature set $\mathcal{F}$ of length $L$ with
elements as $\{f_i\}$ $(i\in 1,2,\cdots,L)$.

\section{Feature Selection} \label{sec:mrmr}
In order to select the relevant features and to eliminate the
redundant features, the mRMR feature selection technique is used
\cite{peng2005feature}. The feature selection process also reduces
the length of feature vectors and thus lowers the effective
computational complexity. To calculate mRMR ranking of a dynamic
feature $f_i(t)$ $(i\in 1,2,\cdots,L)$, we employ the difference
between the maximum relevance and minimum redundancy criteria
expressed in terms of their dynamic random variables $F_i(t)$
$(i\in 1,2,\cdots,L)$ given by \cite{peng2005feature}
\begin{align}
\Psi_{R\mathcal{F}}(t)=
\max_{F_i\in\mathcal{F}}\bigg[\frac{1}{|\mathcal{F}|}&\sum_{F_i\in\mathcal{F}}\mathcal{M}(R(t),F_i(t))-\nonumber\\
&\frac{1}{|\mathcal{F}|^2}\sum_{F_i,F_j\in\mathcal{F}}\mathcal{M}(F_i(t),F_j(t))\bigg]
\end{align}
where $R(t)$ is the dynamic random variable of the ground truth of
instantaneous emotional rating $r(t)$, $\mathcal{M}(\cdot)$ is the
mutual information of two random variables $F_i$ and $F_j$ given
by
\begin{align}
\mathcal{M}(F_i,F_j)=\int\int
p(F_i,F_j)\log\frac{p(F_i,F_j)}{p(F_i)p(F_j)}\textrm{d}F_i\textrm{d}F_j
\end{align}
where $p(F_i)$, $p(F_j)$ and $p(F_i,F_j)$ are the probability
density functions. Finally, the feature vector $\bm{\textbf{F}_s}$
consisting only the features $\{f_{si}\}$ $(i\in 1,2,\cdots,L_s)$
with high values of mRMR ranking are constructed to predict the
emotional states.

\section{Prediction of Emotional Rating}
In the proposed method, a regression technique is required to map
the features to continuous-level emotional dimension. We employ
the support vector regression (SVR)~\cite{svr1997} technique to
predict the emotional states from the proposed audiovisual
features by acknowledging that it is a well-established
statistical learning theory applied successfully in many
prediction tasks in computer vision. The kernel SVR implicitly
maps the dynamic features into a higher dimensional feature space
to find a linear hyperplane, wherein the emotional state can be
predicted with a predefined soft error margin.

Given a training set of known emotional rating
$\bm{\Theta}(t)\in\left\{\bm{\textbf{F}_{s}}(t),r(t)\right\}$,
where $\bm{\textbf{F}_{s}}(t)\in \mathbb{R}^{L_s}$ and $-1\leq
r(t)\leq 1$, the emotional state is predicted using the test
feature $\bm{\hat{\textbf{F}}_{s}}(t)$ as a regression function
given by
\begin{align}
\hat{r}(t)=\sum_{i=1}^{L_s}\beta_i
\Phi\left(f_{si}(t),\hat{f}_{si}(t)\right)+b
\end{align}
where $\beta_i$ are the Lagrange multipliers of a dual
optimization problem, $\Phi\left(\cdot\right)$ is a kernel
function, $f_{si}$ are the support vectors, and $b$ is the weight
of bias. In order to map the audiovisual features into the higher
dimensional feature space for prediction, the most frequently used
kernel functions such as the linear, polynomial, and radial basis
function (RBF) can be used. With a view to select the parameters
of the SVR, a grid-search on the hyper-parameters is used by
adopting a cross-validation scheme. The parameter settings that
produce the best cross-validation accuracy are used for predicting
the emotional state from the proposed CNN-based features.

\section{Experimental Results}
Experiments are carried out to evaluate the performance of the
prediction method using features selected from the extracted
features of the proposed CNNs on a multimodal corpus of
spontaneous interactions in French, called the REmote
COLlaborative and Affective interactions (RECOLA), introduced
in~\cite{ringeval2013introducing}. The RECOLA database includes
9.5 hours of multimodal recordings such as the audio, video,
electro-cardiogram and electrodermal activities that were
continuously and synchronously recorded from $46$ participants.
Time- and value-continuous ratings of emotion were performed by
six French-speaking assistants (three male, three female) for the
first five minutes of all recorded sequences.
% The dataset for which
%the participants gave their consents to share their data is reduced to a
%set of $34$ participants for an overall duration of seven hours.
Finally, the annotations of $10$ male and $13$ female participants
were made publicly available.
%The data contains annotations for $10$ male and $13$ female
%participants with mean age of $21.3$ years and standard deviation
%of $4.1$ years. The participants were French speakers, however, they
%were originated from different mother tongues: $17$ subjects were French, three German
%and three Italian.
In the experiments, we used only the audio and video modalities
from the database for predicting the continuous-level emotional
scores. Among the subjects, videos of first $10$ subjects have
been chosen for training the CNN model and the videos of the
remaining subjects have been chosen for testing purposes.

\begin{table}[t]
\centering \caption{Comparison of prediction performance of
emotional dimension valance in terms of RMSE, CC and CCC using
different features}\label{tab:result}
\renewcommand{\arraystretch}{1.25}
\begin{small}
\begin{tabular}{|c|c|c|c|}
\hline Features & RMSE & CC & CCC \\
\hline Audio (MFCC) & $0.0696$ & $0.5582$ & $0.0069$ \\
\hline Visual (LBP-TOP) & $0.0472$ & $0.8032$ & $0.1137$ \\
\hline Visual~\cite{dahmane2011continuous} (Gabor Energy) & $0.0481$ & $0.8440$ & $0.0431$ \\
\hline \multirow{2}{*}{Audiovisual~\cite{sudipta2016affect}
\vspace{0.35cm}} & \multirow{3}{*}{$0.0469$} & \multirow{3}{*}{$0.7998$} & \multirow{3}{*}{$0.2592$} \\
(MFCC with $\delta$ and $\delta\delta$ & & & \\
+ LBP-TOP) & & & \\
\hline \multirow{2}{*}{Proposed Audiovisual \vspace{0.25cm}} & \multirow{2}{*}{$\mathbf{0.0364}$} & \multirow{2}{*}{$\mathbf{0.8924}$} & \multirow{2}{*}{$\mathbf{0.3668}$}\\
(CNN) & & & \\
\hline
\end{tabular}
\end{small}
\end{table}

\begin{figure*}[t]
\centering
\includegraphics[width=8cm,height=5.5cm]{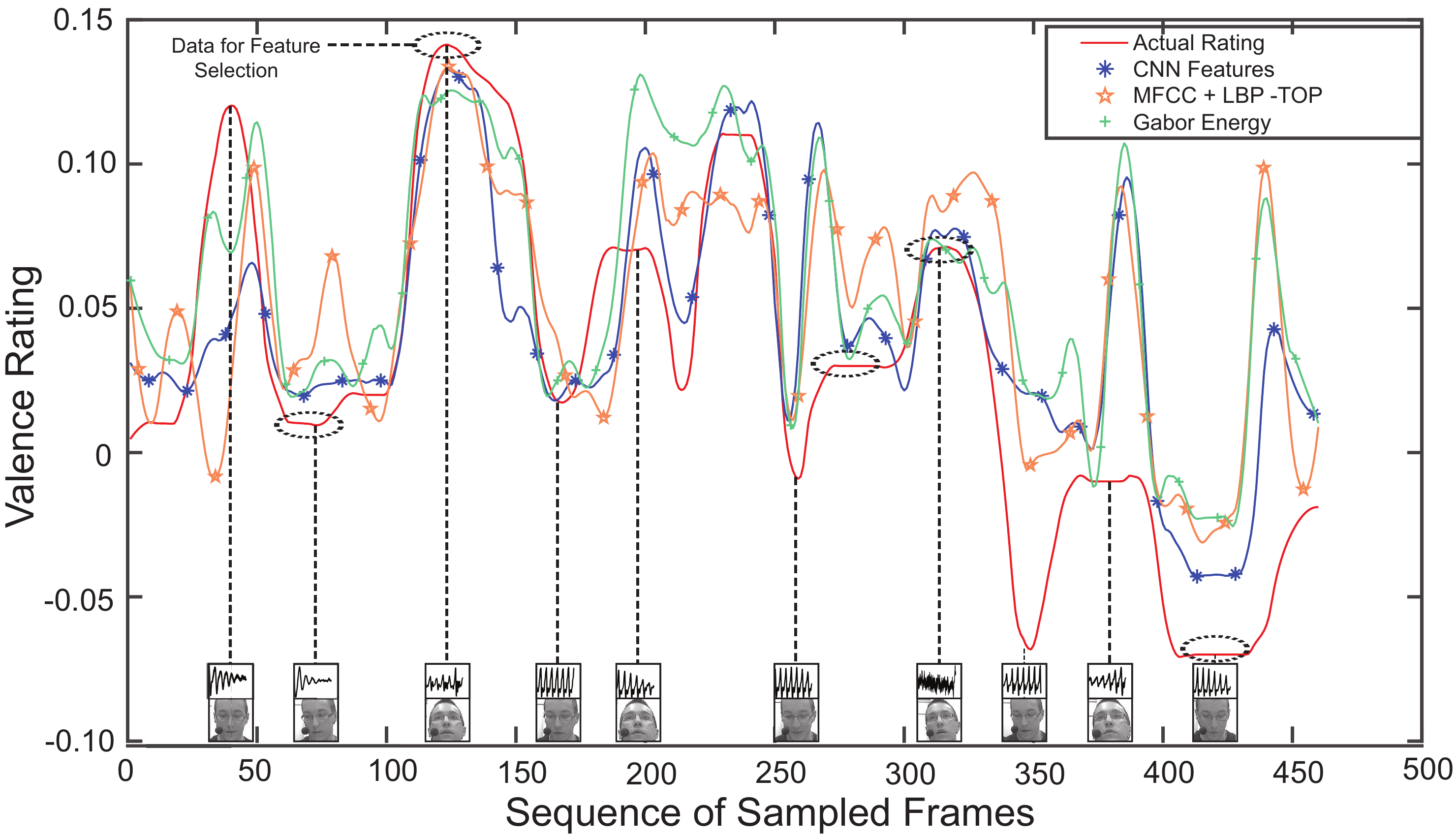}\hspace{0.25cm}
\includegraphics[width=8cm,height=5.5cm]{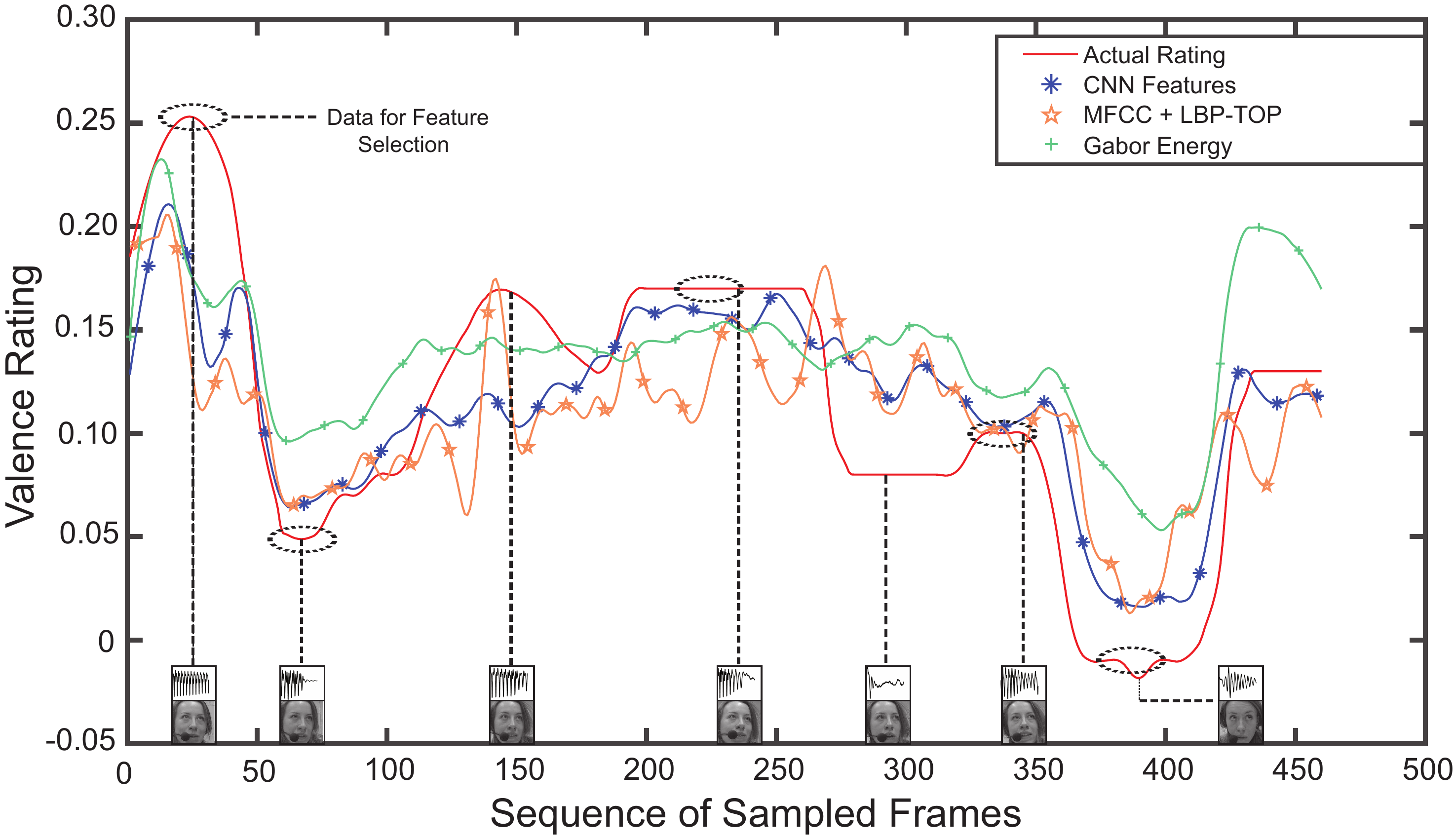}\\
(a)\hspace{7cm}(b)\caption{Prediction of valance dimension using
the proposed method for subject number (a) $16$ and (b) $23$ of
RECOLA database.}\label{fig:tracking}
\end{figure*}

The input of the CNN for video stream is considered to be
$128\times128$ pixels greyscale video frames. The number of
filters in the weight vectors $\mathbf{W_1^{(v)}}$,
$\mathbf{W_2^{(v)}}$, $\mathbf{W_3^{(v)}}$ and
$\mathbf{W_r^{(v)}}$ with corresponding bias terms
$\mathbf{b_1^{(v)}}$, $\mathbf{b_2^{(v)}}$, $\mathbf{b_3^{(v)}}$
and $b_r^{(v)}$, are set to $128$, $256$, $512$ and $1$,
respectively. The kernel size of both of the convolutional weights
$\mathbf{W_1^{(v)}}$ and $\mathbf{W_2^{(v)}}$ is set to
$5\times5$.
%The weights
%$\mathbf{W_1^{(v)}}$ includes $128$ filters and $\mathbf{W_1^{(v)}}$
%includes $256$ filters with kernel size $5\times5$, and stride $1$ in
%both weights. The output of the second stage is $32\times32\times256=262144$
%dimensional. To map the features in a $512$ dimensional array, the kernel
%size of the weights $\mathbf{W_3^{(v)}}$ of the fully connected operation
%of the third stage is set to $262144\times512$. The final weight vector
%$\mathbf{W_r^{(v)}}$ of the regressor to estimate the emotional score is
%thus set to $512\times1$. The dimensions of the bias terms are
%$128$, $256$, $512$ and $1$, respectively.
The CNN for audio stream has been set up in a similar fashion. The
input audio stream of each frame is taken to be of $60$ ms in
length, with a $20$ ms overlap with audio stream of previous
frame, sampled at $44.1$ KHz. Thus,
%the input of the CNN is $2646$ dimensional audio stream. The weight vector
the number of filters in the weight vectors $\mathbf{W_1^{(a)}}$,
$\mathbf{W_2^{(a)}}$, $\mathbf{W_3^{(a)}}$ and
$\mathbf{W_r^{(a)}}$ with corresponding bias terms
$\mathbf{b_1^{(a)}}$, $\mathbf{b_2^{(a)}}$, $\mathbf{b_3^{(a)}}$
and $b_r^{(a)}$ are set to $32$, $64$, $512$ and $1$,
respectively. The kernel size of both of convolutional weights
$\mathbf{W_1^{(a)}}$ and $\mathbf{W_2^{(a)}}$ is chosen to be $20$
with stride $2$.

The feature vector $\mathcal{F}$ with length $1024$ is constructed
by concatenating learned affective features $\mathbf{X_3^{(v)}}$
and $\mathbf{X_3^{(a)}}$, each with $512$ dimensions. The
effective features are selected from this feature vector using the
mRMR pipeline as explained in Section~\ref{sec:mrmr}.
Approximately $25\%$ frames of video clips are considered for
selection of effective features and the rest of the frames for
predicting the emotional dimensions. We have discretized the range
of ratings in $10$-levels uniformly. The sets of $50$-neighboring
frames that have variation less than $20\%$ for each of the levels
are chosen for the purpose of selection process. The parameters
such as the length of RBF kernel, weights, and bias of the SVR are
optimized in terms of the mean absolute error (MAE) with five fold
cross validation. The optimized SVR is employed first to select
effective length of features those are ranked by the criterion
mRMR, and finally for prediction of emotional dimension on the
testing frames of the video.

% \subsection{Results}

The  overall  performance  of  prediction  is  compared with
proposed learned audiovisual features from CNN, as well as audio
features MFCC, visual features including  LBP-TOP, the  Gabor
energy~\cite{dahmane2011continuous}, and Paul's ensemble of MFCC
with $\delta$ and $\delta\delta$ and
LBP-TOP~\cite{sudipta2016affect}. Table~\ref{tab:result} shows the
overall prediction performance of the testing clips in terms of
root mean squared error (RMSE), Pearson's correlation coefficient
(CC), and Lin's concordance correlation coefficient (CCC). It is
seen from the table that the proposed method of using audiovisual
features learned by CNNs shows a $28.8\%$ improvement of RMSE from
the nearest method that employs hand-crafted features. Moreover,
proposed method shows a $5.4\%$ improvement of CC and $29.3\%$
improvement of CCC from the method showing closest performance.

Figure~\ref{fig:tracking} shows instantaneous prediction of the
emotional dimension valence for subject $16$ (male) and subject
$23$ (female) of RECOLA database. The regions chosen for feature
selection process are marked in the figures. It can be observed
from the figures that the proposed method can closely follow the
affective ratings of ground truth. Moreover, it can be observed
that Paul's method as well as the Gabor energy curves shows sudden
changes in estimation, which is in the complete opposite direction
of the trend of the ground truth. Such, a misleading trajectory is
absent for the case of proposed CNN-based features, which ensures
that the learned features of CNN are robust. Also, if there is a
sudden change of a dimension, then the proposed prediction can
follow the trend within few seconds as per the frame-rate. Thus,
the proposed instantaneous emotion prediction technique can be
effective in developing real-time sensitive artificial listener
(SAL) agents.

\section{Conclusion}
Automatic prediction of emotional states is crucial for developing
SAL that has many potential applications requiring interaction
between machines and humans. The generalized approach of
prediction of emotional state follows the steps of extraction of
affective features, selection of features, and mapping of selected
features using a regressor. This paper has investigated the
performance of instantaneous prediction of commonly-referred
emotional dimensions, such as valence, using the extracted
audiovisual features learned by CNNs. Extracted features are then
selected by using the mutual information-based mRMR ranking. These
low-level features are mapped on the emotional dimensions using
the SVR technique. Performance of the proposed audiovisual
features is compared with existing audio and visual features for
prediction of instantaneous rating of emotional dimensions. The
RMSE, CC and CCC calculated using different types of features show
that the prediction performance improves significantly, when top
ranked features are considered for the regression. Experiments on
instantaneous prediction reveal that a moderate length audiovisual
features learned by the proposed CNN-based method presented in
this paper can provide a few seconds of settling time even when an
emotional dimension changes sharply.

\bibliographystyle{IEEEtran}

%%%%\bibliography{ramesh_aff}

\end{document}